\documentclass[a4paper]{article}
\usepackage{iwslt17}
\usepackage{times}
\usepackage{url}
\usepackage{latexsym}
\usepackage{bm}
\usepackage{amssymb}
\usepackage{epsfig}
\usepackage{amsmath}
\usepackage{multirow}
\usepackage{graphicx}
\usepackage{caption}
\usepackage{float}
\usepackage{subcaption}
\usepackage{dblfloatfix}
\usepackage[T1]{fontenc}
\usepackage[utf8]{inputenc}
\usepackage[titletoc,toc,title]{appendix}
\def\reg{{\rm\ooalign{\hfil
     \raise.07ex\hbox{\scriptsize R}\hfil\crcr\mathhexbox20D}}}

\newcommand{\todo}[1]{}
\renewcommand{\todo}[1]{{\color{red} TODO: {#1}}}

\newcommand{\dotproduct}{\raisebox{.3ex}{\tiny$\bullet$}} 

\title{Effective Strategies in Zero-Shot Neural Machine Translation}

 \makeatletter
 \def\name#1{\gdef\@name{#1\\}}
 \makeatother
 \name{
  {\em Thanh-Le Ha, Jan Niehues, Alexander Waibel}
 }

\address{Institute for Anthropomatics and Robotics \\
KIT - Karlsruhe Institute of Technology, Germany \\
{\small \tt firstname.lastname@kit.edu} \\
}

\date{}

\begin{document}
\maketitle
\begin{abstract}
In this paper, we proposed two strategies which can be applied to a multilingual neural machine translation 
system in order to better tackle zero-shot scenarios despite not having any parallel corpus. 
The experiments show that they are effective in terms of both performance and computing resources, especially in multilingual translation of unbalanced data in real zero-resourced condition when they alleviate the language bias problem. 
\end{abstract}

\section{Introduction}
\label{intro}
The newly proposed neural machine translation~\cite{Bahdanau2014} has shown the best performance in recent machine translation campaigns for several language pair. 
Being applied to multilingual settings, neural machine translation (NMT) systems have been proved to be benefited from additional information 
embedded in a common semantic space across languages. 
However, in the extreme cases where no parallel data is available to train such system, often NMT systems suffer a bad training situation 
and are incapable to perform adequate translation. 

In this work, we point out the underlying problem of current multilingual NMT systems when dealing with zero-resource scenarios. 
Then we propose two simple strategies to reduce adverse impact of the problem. 
The strategies need little modifications in the standard NMT framework, yet they are still able to achieve 
better performance on  zero-shot translation tasks with much less training time.

\subsection{Neural Machine Translation}
In this section, we briefly describe the framework of Neural Machine Translation as a sequence-to-sequence modeling problem following the proposed method of ~\cite{Bahdanau2014}.

Given a source sentence $\bm{x} = (x_1, .., x_i, .., x_I)$ and the corresponding target sentence $\bm{y} = (y_1, .. , y_j, .., y_J)$, 
the NMT aims to directly model the translation probability of the target sequence:

$$ P(\bm{y}|\bm{x}) = \prod_{j=1}^I P(y_j|y_{<j}, \bm{x}; \bm{\theta}) \nonumber $$

\cite{Bahdanau2014} proposed an \textit{encoder-attention-decoder} framework to calculate this probability.

A bidirectional recurrent \textit{encoder} reads a word $x_{i}$ from the source sentence and produces a representation of the sentence in a fixed-length vector $\bm{h}_i$ 
concatenated from those of the forward and backward directions:
%\begin{equation}
\begin{align}
& \bm{h}_i=[\overrightarrow{\bm{h}}_i,\overleftarrow{\bm{h}}_i] \nonumber \\
& \overrightarrow{\bm{h}}_i=d(\overrightarrow{\bm{h}}_{i-1},\bm{E}_s~\dotproduct~\bm{x}_i) \label{eq1} \\
& \overleftarrow{\bm{h}}_i=d(\overleftarrow{\bm{h}}_{i+1},\bm{E}_s~\dotproduct~\bm{x}_i) \label{eq2} 
\end{align}
%\end{equation}

where $\bm{E}_s$ is the source word embedding matrix to be shared across the source words $x_{i} \in V_x$, $d$ is the recurrent unit 
computing the current hidden state of the encoder based on the previous hidden state. 
$\bm{h}_i$ is then called an \textit{annotation vector} which encodes the source sentence up to the time $i$ from both forward and backward directions. 

Then an \textit{attention mechanism} is set up in order to choose which annotation vectors should contribute to the predicting decision of the next target word. 
Normally, a relevance score $rel(\bm{z}_{j-1},\bm{h}_i)$ between the previous target word and the annotation vectors is used to calculate the context vector $\bm{c}_i$:

 \begin{table*}[t]
  \begin{center}
   \begin{tabular}{c|c|l} \hline \hline
\multicolumn{3}{c}{\textbf{Original corpus}} \\ \hline
Source Sentence 1 & De & versetzen Sie sich mal in meine Lage ! \\
Target Sentence 1 & En & put yourselves in my position .  \\
Source Sentence 2 & En & I flew on Air Force Two for eight years . \\
Target Sentence 2 & Nl & ik heb acht jaar lang met de Air Force Two gevlogen .  \\ \hline
\multicolumn{3}{c}{\textbf{Preprocessed by \cite{Ha2016}}} \\ \hline
Source Sentence 1 & De & \textbf{<en>} \textbf{<en>} \textit{de\_}versetzen \textit{de\_}Sie \textit{de\_}sich \textit{de\_}mal \textit{de\_}in \textit{de\_}meine \textit{de\_}Lage \textit{de\_}! \textbf{<en>} \textbf{<en>}\\
Target Sentence 1 & En & \textit{en\_\_} \textit{en\_}put \textit{en\_}yourselves \textit{en\_}in \textit{en\_}my \textit{en\_}position \textit{en\_}.  \\
Source Sentence 2 & En & \textbf{<nl>} \textbf{<nl>} \textit{en\_}I \textit{en\_}flew \textit{en\_}on \textit{en\_}Air \textit{en\_}Force \textit{en\_}Two \textit{en\_}for \textit{en\_}eight \textit{en\_}years \textit{en\_}. \textbf{<nl>} \textbf{<nl>}\\
Target Sentence 2 & Nl & \textit{nl\_\_} \textit{nl\_}ik \textit{nl\_}heb \textit{nl\_}acht \textit{nl\_}jaar \textit{nl\_}lang \textit{nl\_}met \textit{nl\_}de \textit{nl\_}Air \textit{nl\_}Force \textit{nl\_}Two \textit{nl\_}gevlogen \textit{nl\_}.  \\ \hline
\multicolumn{3}{c}{\textbf{Preprocessed by \cite{Johnson2016}}} \\ \hline
Source Sentence 1 & De & \textbf{2en} versetzen Sie sich mal in meine Lage ! \\
Target Sentence 1 & En & put yourselves in my position .  \\
Source Sentence 2 & En & \textbf{2nl} I flew on Air Force Two for eight years . \\
Target Sentence 2 & Nl & ik heb acht jaar lang met de Air Force Two gevlogen . \\
 \hline \hline
   \end{tabular}
 \caption{\label{Example} Examples of preprocessing steps conducted by \cite{Ha2016} and \cite{Johnson2016}.}
  \end{center}
  \vspace*{0.2cm}
 \end{table*}

\[
\begin{aligned}
& \alpha_{ij} = \displaystyle  \frac{\exp(rel(\bm{z}_{j-1},\bm{h}_i))}{\sum_{i'} \exp(rel(\bm{z}_{j-1},\bm{h}_{i'}))},~\bm{c}_j = \displaystyle \sum_{i}{\alpha_{ij}\bm{h}_i}\\
\end{aligned}
\]
 
In the other end, a \textit{decoder} recursively generates one target word $y_{j}$ at a time:
\begin{equation}
P(y_j|y_{<j}, \bm{x}; \bm{\theta}) = \frac{\exp{(\bm{z}_{j})}}{\sum_{k=1}^{|V_y|} \exp{(\bm{z}_{k})}} \nonumber
\end{equation}

Where: 
\begin{align}
& \bm{z}_{j}=g(\bm{z}_{j-1}, \bm{t}_{j-1}, \bm{c}_j) \nonumber \\
& \bm{t}_{j-1} = \bm{E}_t~\dotproduct~\bm{y}_{j-1} \label{eq3} 
\end{align}

The mechanism in the decoder is similar to its counterpart in the encoder, excepts that beside the previous hidden state $\bm{z}_{j-1}$ 
and target embedding $\bm{t}_{j-1}$, it also takes the context vector $\bm{c}_j$ from the attention layer as inputs to calculate the current hidden state $\bm{z}_{j}$. 
The predicted word $y_j$ at time $j$  then can be sampled from a softmax distribution of the hidden state. 
Basically, a beam search is utilized to generate the output sequence - the translated sentence in this case.

\subsection{Multilingual NMT}
\label{multiling}
State-of-the-art NMT systems have demonstrated that machine translation in many languages can achieve high quality results with large-scale data and sufficient computational
power\cite{bojar2016findings,cettolo2016iwslt}. On the other hand, how to prepare such enormous corpora for low-resourced languages and specific domains has remained a big problem. 
Especially in zero-resourced condition where we do not possess any bilingual corpus, building a data-driven translation system requires special techniques that can enable some sort of transfer learning.
A simple but effective approach called pivot-based machine translation has been developed. 
The idea of the pivot-based approach is to indirectly learn the translation of the source and target languages through a bridge language. 
However, this pivot approach is not ideal since it is necessary to build two different translation systems for each language pair in order to perform the bridge translation, 
hence possibly produces more ambiguities cross languages as well as error-prone to the individual systems.

\begin{figure*}[t]
\begin{center}
\hspace*{-2cm}
\includegraphics[scale=0.5]{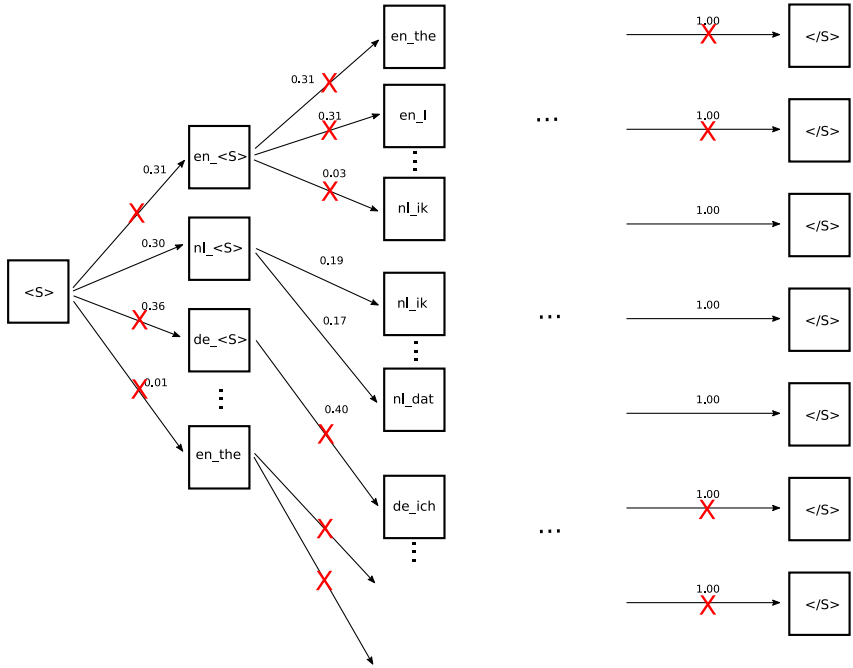}
\caption{\label{fig:filtereddict} Effect of target dictionary filtering on the decoding process using beam search.}
\end{center}
  \vspace*{0.3cm}
\end{figure*} 

Recent work has started exploring potential solutions to perform machine translation for multiple language pairs using a single NMT system.
One of the most notable differences of NMT compared to the conventional statistical approach is that the source words can be represented in a continuous space 
in which the semantic regularities are induced automatically. 
Being applied to multilingual settings, NMT systems have been proved to be benefited from additional information embedded in a common semantic space across languages, 
thus, by some means they are able to conduct some level of transfer learning. 

In this section, we review the related work on constructing a multilingual NMT system involved in translating from several source languages to several target languages. 
Then we consider a potential application of such a multilingual system on zero-shot scenarios to demonstrate the capability of those systems in extreme low-resourced conditions. 

We can essentially divided the work into two directions in applying the current NMT framework for multilingual scenarios. 
The first direction follows the idea that multilingual training of an NMT system can be seen as a special form of multi-task learning 
where each encoder is responsible to learn an individual modality's representation and each decoder's mission is to predict labels of a particular task. 
In such a multilingual system, each task or modality corresponds to a language. 
In \cite{Luong2016}, the authors utilizes a multiple encoder-decoder architecture to do multi-task learning, including many-to-many translation, parsing and image captioning. 
\cite{Firat2016} proposed another approach which enable attention-based NMT to multilingual translation.  
Similar to \cite{Luong2016}, they use one encoder per source language and one decoder per target language for  \textit{many-to-many} translation tasks. 
Instead of a quadratic number of independent attention layers, however, their NMT system contains only a single, huge attention layer. 
In order to achieve this, the attention layer need to be provided some sort of aggregation layer between it and the encoders as well as the decoders. 
It is required to change their architecture to accommodate such a complicated shared attention mechanism. 

The work along the second direction also considers multilingual translation as multi-task learning, although the tasks should be the same (i.e. translation) 
with the same modality (i.e. textual data). The only difference here is whether we decide which components are shared across languages or we let the architecture learns to share what.
In \cite{quan2017kit}, the authors developed a general framework to analyze which components should be shared in order to achieve the best multilingual translation system. 
Other works chose to share every components by grouping all language vocabularies into a large vocabulary, 
then use a single encoder-decoder NMT system to perform many-to-many translation as each word is viewed as a distinct entry  in the large vocabulary regardless of its language. 
By implementing such mechanism in the preprocessing step, those approaches require little or no modification in the standard NMT architecture.
In our previous work\cite{Ha2016}, we performed a two-step preprocessing:
\begin{enumerate}
 \item \textit{Language Coding}: Add the language codes to every word in source and target sentences.
 \item \textbf{Target Forcing}: Add a special token in the beginning of every source sentence indicating the language they want the system to translate the source sentence to.\footnote{In fact,
  we add the target language token both to the beginning and to the end of every source sentence, each place two times, to make the forcing effect stronger. Furthermore, every target sentence starts with 
  a pseudo word playing the role of a start token in a specific target language. This pseudo word is later removed along with sub-word tags in post-processing steps.}
\end{enumerate}
Concurrently, \cite{Johnson2016} proposed a similar but simpler approach: they carried out only the second step as in the work of \cite{Ha2016}. 
They expected that there would be only a few cases where two words in difference languages (with different meanings) having the same surface form. 
Thus, they did not conduct the first step. An interesting side-effect of not doing language-code adding, as \cite{Johnson2016} suggested, 
is that their system could accomplish code-switching multilingual translation, i.e. it could translate a sentence containing words in different languages.
The main drawback of these approaches is that the sizes of the vocabularies and corpus grow proportionally to the number of languages involved. 
Hence, a huge amount of time and memory are necessary to train such a multilingual system. Table~\ref{Example} gives us a simple example illustrating those preprocessing steps.

\vspace*{0.2cm}
\section{Multilingual-based Zero-Shot Translation}
\label{proposed}
In this section, we follow the second direction of \cite{Ha2016} and \cite{Johnson2016}, hereby called \textit{mix-language} approaches. 
First we built some baselines inspired of their approaches and participated in the new challenge of zero-shot translation at IWSLT 2017. 
Then we proposed two strategies, \textit{filtered dictionary} and \textit{language as a word feature}, in attempts to tackle the drawbacks of their approaches. 
The results in section~\ref{results} show that our strategies are highly  effective in terms of both performance and training resources.

\subsection{Target Dictionary Filtering}
In \cite{Ha2016}, the authors discussed about observations of the language bias problem in our multilingual system: If the very first word is wrongly translated into wrong language, 
the following picked words are more probable in that wrong language again. The problem is more severe when the mixed target vocabulary is unbalanced, 
due to the language unbalance of the training corpora (whereas the zero-shot is a typical example). 
We reported a number of 9.7\% of the sentences wrongly translated in our basic zero-shot German$\rightarrow$French system. 

One solution for this problem is to enhance the balance of the corpus by adding \textit{target}$\rightarrow$\textit{target} corpora into the multilingual system as suggested in \cite{Ha2016}. 
The beam search still need to consider, however, other candidates belonging to the target vocabulary that should not be considered. 
In this work, we propose a simple yet effective technique to eliminate this bad effect. 
In the translation process to a specific language,
we filter out all the entries in the languages other than that desired language from the target vocabulary. 
It would significantly reduce the translation time in huge multilingual systems or big texts to be translated due to the fact that 
many search paths containing the unwanted candidates are removed. More importantly, it assures the translated words and sentences are in the correct language. 
The effect of this strategy in the decoding process is illustrated in Figure~\ref{fig:filtereddict}.

\begin{table*} [t] 
\label{table:underresourced}
\centerline{ 
\begin{tabular}{|c|l|c|c|c|c|c|}
\hline \hline
 & \multirow{2}{*}{\textbf{System}} & \multirow{2}{*}{\textbf{Zero-shot?}}  & \multicolumn{2}{c|}{ \textbf{German$\rightarrow$Dutch}} & \multicolumn{2}{c|}{ \textbf{German$\rightarrow$Romanian}}  \\ 
\cline{4-7}
 & & &
dev2010 & tst2010 & dev2010 & tst2010   \\ \hline
(1) & Direct & No & 17.83 & 20.49 & 12.41 & 15.14 \\
(2) & Pivot (via English) & Yes & 16.11 & 19.12 & 12.88 & 15.04 \\
\hline
(3) & Zero 2L \cite{Johnson2016} & Yes & 4.79 & 5.75 & 1.55 & 2.05   \\ 
(4) & Zero 4L \cite{Johnson2016} & Yes & 6.31 & 7.93 & 3.15 & 3.73  \\ \hline
(5) & Zero 6L \cite{Ha2016} & Yes & 11.58 & 14.95 & 8.61 & 10.83  \\
(6) & Back-Trans \cite{Ha2016} & No & 17.33 & 20.36 & 12.92 & 15.62  \\ 
\hline \hline
\end{tabular}}
\caption{\label{table:general} {Results of the popular \textit{mix-language} methods applied to German$\rightarrow$Dutch 
and German$\rightarrow$Romanian zero-shot tasks.}}
\vspace*{0.2cm}
\end{table*}

\subsection{Language as a Word Feature}
As briefly mentioned in Section~\ref{multiling}, the main disadvantage of the \textit{mix-language approaches} is the efficiency of training process. 
Usually in those systems, source and target vocabularies have a huge number of entries, in proportion to the number of languages whose corpora are mixed. 
It leads to immerse numbers of parameters laying between the embedding and hidden states of the encoder and the decoder. 
More problematic is the size of the output softmax - where most calculations take place. 

There exist works on integrating linguistic information into NMT systems in order to help predict the output words\cite{SennrichH16,hoang2016improving,niehues2016using}. 
In those works, the information of a word (e.g. its lemma or its part-of-speech tag) are integrated as a word features. 
It is conducted simply by learning the feature embeddings instead of the word embeddings.
In other words, their system considers a word as a special feature together with other features of itself. 

\begin{figure}[h]
\begin{center}
\vspace*{1cm}
%\hspace*{-2cm}
\includegraphics[scale=0.4]{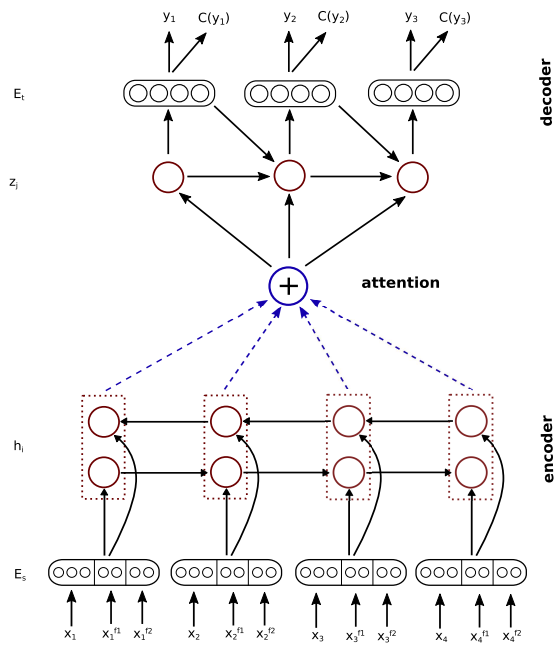}
\caption{\label{fig:langfeature} The NMT architecture which allows the integration of linguistic information as word features.}
\end{center}
  \vspace*{-0.3cm}
\end{figure}

More specially, in the formula~\ref{eq1},~\ref{eq2} and ~\ref{eq3}, the embedding matrices are the concatenation of all features' embeddings:
\[
\begin{aligned}
\bm{E} ~\dotproduct~\bm{x} = \displaystyle \underset{f\in F}{\Big[,\Big]}(\bm{E}^f~\dotproduct~\bm{x}^f)
\end{aligned}
\]
Where $\Big[,\Big]$ is the vector concatenation operation, concatenating the embeddings of individual feature $f$ in a finite, arbitrary set $F$ of word features. 
The target features of each target word would be jointly predicted along the word. Figure~\ref{fig:langfeature} denotes this modified architecture.

Inspired by their work, we attempt to encode the language information directly in the architecture instead of performing language token attachment in the preprocessing step. 
Being applied in our model, instead of the linguistic information at the word level, 
our source word features are the language of the considering word and the correct language the target sentence
The only target feature is the language of the produced word by the system. 
For example, when we would like to translate from the sentence ``\textit{put yourselves in my position}'' into German,
the features of each source word would be the word itself, e.g. ``\textit{yourselves}'', and two additional features ``\textit{en}'' and ``\textit{de}''. 
Similarly, the features of the target words are the word and ``\textit{de}''. 
This scheme of using language information looks alike to \cite{Ha2016}, but the difference is the way the language information are integrated into the NMT framework. 
In \cite{Ha2016}, those information are implicitly injected into the system. In this work, they are explicitly provided along with the corresponding words. 
Furthermore, when being used together in the embedding layers, they can share useful information and constraints 
which would be more helpful in choosing both correct words and language to be translated to.
During decoding, the beam search is only conducted on the target words space and not on the target features. 
When the search is complete, the corresponding features are selected along the search path. 
In our case, we do not need the output of the target language features excepts for the evaluation of language identification purpose.
\vspace*{0.2cm}
\section{Evaluation}
\label{evaluation}
In this section, we describe a thorough evaluation of the related methods in comparisons with the direct approach as well as the pivot-based approach. 
\subsection{Experimental Settings}
\label{settings}
We participated to this year's IWSLT zero-shot tasks for German$\rightarrow$Dutch and German$\rightarrow$Romanian. 
The pivot language used in our experiments is English and the parallel corpora are German-English and English-Dutch or German-English and English-Romanian.
The data are extracted from WIT3's\footnote{\url{https://wit3.fbk.eu/}} TED corpus\cite{cettolo2012}. 
The validation and test sets are {\tt dev2010} and {\tt tst2010} which are provided by the IWSLT17 organizers.

We use the Lua version of {\tt OpenNMT}\footnote{\url{http://opennmt.net/}}\cite{klein2017opennmt} framework to conduct all experiments in this paper.
Subword segmentation is performed using Byte-Pair Encoding \cite{Sennrich2016a} with 40000 merging operations.
All sentence pairs in training and validation data which exceeds 50-word length are removed and the rest are shuffled inside each of every minibatch.
We use 1024-cell LSTM layers\cite{HochreiterLSTM} and 1024-dimensional embeddings with dropout of 0.3 at every recurrent layers. 
The systems are trained using Adam\cite{kingma2014adam}. 
In decoding process, we use a beam search with the size of 15.
\vspace*{0.1cm}
\subsection{Baseline Systems}
\label{baseline}
Let us consider the scenario that we would like to translate from a \textit{source} language to a \textit{target} language via a \textit{pivot} language. 
In order to evaluate the effectiveness of our proposed strategies, we reimplemented the following baseline systems: 
\begin{itemize}
 \item \textit{Direct}: A system which does not exist in the real world is trained using the parallel corpus. It is only for comparison purpose.
 \item \textit{Pivot}: A system which uses English as the pivot language. The output of the first \textit{source}$\rightarrow$\textit{pivot} translation system 
 was pipelined into the second system trained to translate from \textit{pivot} to \textit{target}.
 \item \textit{Zero 2L}: To build this system, we followed the idea of \cite{Johnson2016}: we added a target token to every \textit{source} sentences in the parallel corpus of \textit{source}$\rightarrow$\textit{pivot}, 
 added another target token to every \textit{pivot} sentences in the parallel corpus of \textit{pivot}$\rightarrow$\textit{target}, 
 merged those two parallel corpora into a big corpus and used our standard NMT architecture mentioned in previous section to train and decode. 
 The only differences are the actual data and a simpler NMT architecture we used to train the system.
 \item \textit{Zero 4L}: Same as \textit{Zero 2L} but in addition applying 
 to two other directions \textit{pivot}$\rightarrow$\textit{source} and \textit{target}$\rightarrow$\textit{pivot}. 
 The result is a parallel corpus two times larger than the corpus in \textit{Zero 2L}.
 \item \textit{Zero 6L}: This is an extended version of our previous work\cite{Ha2016}. There are two main differences compared \textit{Zero 2L} and \textit{Zero 4L}: we conducted both Language Coding and Target Forcing preprocessing steps, 
 the data used to trained are actually six parallel corpora: \textit{source}$\leftrightarrow$\textit{pivot}, \textit{pivot}$\rightarrow$\textit{pivot},
 \textit{pivot}$\leftrightarrow$\textit{target}, \textit{target}$\rightarrow$\textit{target}. Finallly we merged them at the end to form a big parallel corpus.
 \item \textit{Back-Trans}: This is not a real zero-shot system where we back-translated the English part of the \textit{pivot}-\textit{target} parallel corpus
 using a \textit{target}-\textit{pivot} NMT system. 
 At the end we have a \textit{source}-\textit{target} parallel corpus with back-translation quality. 
 After we obtained that direct corpus, we apply the same steps as in the \textit{Zero 6L} setting to all corpora we have (8 parallel corpora in total).
\end{itemize}

%  \textit{Google Model 1} and \textit{Google Model 2} are the realized versions of ~\cite{Johnson2016}. 
% \textit{Zero} is an extended version of \cite{Ha2016}. 
% Compared to their zero-shot configuration, we also including the reverse directions of each parallel corpora. 
% \textit{Zero Back-Trans} is not a real zero-shot system where we back-translate the English part of 
% English-Target parallel corpus using an English$\rightarrow$German NMT system. At the end we have a German-Target parallel corpus with back-translation quality.}
\subsection{Results}
\label{results}
First we applied the baseline systems with respect to the IWSLT17 zero-shot tasks. 
From Table~\ref{table:general} we can see that in general, translating from German$\rightarrow$Romanian is more difficult than German$\rightarrow$Dutch, 
which is reasonable when German and Dutch are considered to be similar. The direct approach which uses a parallel German-\textit{target} corpus and 
the pivot approach have similar performance in term of BLEU score\cite{papineni2002bleu}. 
Interestingly, the \textit{Back-Trans} performed better that the direct approach on German$\rightarrow$Romanian. 
We spectaculate that back translation might pose some translation noise which makes the translation from German$\rightarrow$Romanian more robust.

\begin{table*} [!t] 
\centerline{ 
\begin{tabular}{|c|l|c|c|c|c|c|}
\hline \hline
 & \multirow{2}{*}{\textbf{System}} & \multirow{2}{*}{\textbf{Zero-shot?}}  & \multicolumn{2}{c|}{ \textbf{German$\rightarrow$Dutch}} & \multicolumn{2}{c|}{ \textbf{German$\rightarrow$Romanian}}  \\ 
\cline{4-7}
 & & &
dev2010 & tst2010 & dev2010 & tst2010   \\ \hline
(1) & Zero 2L \cite{Johnson2016} & Yes & 4.79 & 5.75 & 1.55 & 2.05   \\ 
(2) & Zero 4L \cite{Johnson2016} & Yes & 6.31 & 7.93 & 3.15 & 3.73  \\ 
\hline
(3) & Zero 6L \cite{Ha2016} & Yes & 11.58 & 14.95 & 8.61 & 10.83  \\
(3a) & Zero 6L Filtered Dict & Yes & 12.50 & 16.02 & 9.10 & 11.00  \\
(3b) & Zero 6L Lang Feature & Yes & 13.95 & 17.15 & 9.88 & 11.37  \\
\hline
(4) & Back-Trans \cite{Ha2016} & No & 17.33 & 20.36 & 12.92 & 15.62  \\ 
(4a) & Back-Trans Filtered Dict & No & 17.13 & 20.22 & 13.10 & 15.67  \\ 
(4b) & Back-Trans Lang Feature & No & 17.48 & 20.24 & 13.43 & 15.70  \\ 
\hline \hline
\end{tabular}}
\caption{\label{table:strategies} {Effects of the proposed strategies on performance of zero-shot translation systems}}
\end{table*}

Compared to the \textit{Zero 6L} model (5), two other Google-inspired models \textit{Zero 2L} (3) and \textit{Zero 4L} (4) from \cite{Johnson2016} achieved quite low scores. 
This explains the language-bias problem when these models used less and unbalanced corpora than the \textit{Zero 6L} system. 
However, the real zero-shot systems (2, 3, 4, 5), excepts the pivot one (2), performed worse than those using direct parallel corpora (1) and (7),  
since the zero-shot systems have not been shown the direct data, hence, having little or no guide to learn the translation. 
Among those real zero-shot non-pivot systems, the \textit{Zero 6L} system got the best performance due to the amount and the balance of the data used to train. 
Thus, from hereinafter we consider the \textit{Zero 6L} as the baseline to analyze the effectiveness of our proposed strategies.

When we applied the proposed strategies, it is interesting to see their effects on different types of systems. 
Since \textit{Zero 2L} and \textit{Zero 4L} do not have the language identity for words, we cannot directly apply our strategies on those systems. 
In contrast, it is straight-forward to adapt \textit{Target Dictionary Filtering} and \textit{Language as a Word Feature} on the systems described in \cite{Ha2016}.

Table~\ref{table:strategies} shows the performance of our strategies compared to \cite{Ha2016} and \cite{Johnson2016} methods. 
When we applied the strategies on top of \textit{Back-Trans} system, it seems that the data it used to train is sufficient to avoid the language bias problem. 
Thus, our strategies did not have a significant effect of performance on this system (4a vs. 4 and 4b vs. 4).
But on the real zero-shot configuration (3), both strategies helped to improve the systems by notable margins. 
On {\tt tst2010}, \textit{Target Dictionary Filtering} (3a) brought an improvement of 1.07 on German$\rightarrow$Dutch. 
On the same test set, \textit{Language as a Word Feature} achieved the gains of 2.20 BLEU scores compared to \textit{Zero 6L} (3b vs. 3).
On German$\rightarrow$Romanian zero-shot task, the improvements of our strategies were not as great as on German$\rightarrow$Dutch, 
but they still helped, especially on {\tt dev2010}.

Table~\ref{ExampleDict} shows two examples where \textit{Target Dictionary Filtering} clearly improves the quality and readability 
of the translation over the \textit{Zero 6L} when applied.

Considering the effectiveness of our strategy \textit{Language as a Word Feature} on computation perspective, which is shown in Table~\ref{table:time}, 
we observed very positive results. We compared the \textit{Zero 6L} configuration  and our \textit{Language as a Word Feature} system in term of training times,
size of source\&target vocabularies\footnote{In all cases, these sizes are similar numbers.} and the total number of model parameters on both zero-shot translation tasks.
The models were usually trained on the same GPU (Nvidia Titan Xp) for 8 epochs so they are fairly compared (seeing the same dataset the same number of times). 
Each type of models has the same configuration between two zero-shot tasks, 
excepts the parts related to vocabularies\footnote{While the total number of parameters on German$\rightarrow$Romanian 
is bigger than that of German$\rightarrow$Dutch, the training time of German$\rightarrow$Romanian systems is less due to the fact that its training corpus is smaller.}.

By encoding the language information into word features, the number of vocabulary entries reduces to almost half of the original method. 
Thus, it leads to the similar reduction in term of the parameter number. This reduction allows us to use bigger minibatches as well as perform faster updates, 
resulting in substantially decreased training time 
(from 7.3 hours to 1.5 hours for each epoch in case of German$\rightarrow$Dutch and 
from 6.0 hours to 1.3 hours for each epoch in case of German$\rightarrow$Romanian). 
The strategy requires minimum modifications in the standard NMT framework, yet it still achieved better performance with much less training time.

\begin{table} [H] 
\centerline{ 
\begin{tabular}{|l|c|c|c|}
\hline
\multicolumn{4}{|c|}{ \textbf{German$\rightarrow$Dutch}} \\ \hline
\multirow{2}{*}{System} & \#parameters  & Vocab Size & Training Time  \\ 
& (millions)  & (thousands) & (hours$/$epoch)  \\ 
\hline
 Zero 6L & $243$ & $68$ & $7.3$  \\
 Lang Feature & $130$ & $28$ & $1.5$ \\ \hline
\multicolumn{4}{|c|}{ \textbf{German$\rightarrow$Romanian}} \\ \hline
\multirow{2}{*}{System} & \#parameters  & Vocab Size & Training Time  \\ 
& (millions)  & (thousands) & (hours$/$epoch)  \\ 
\hline
 Zero 6L & $247$ & $69$ & $6.0$  \\
 Lang Feature & $122$ & $31$ & $1.3$ \\
 \hline \hline
\end{tabular}}
\caption{\label{table:time} {Effects of the strategy \textit{Language as a Word Feature} on model size and training time.}}
%\vspace*{-0.1cm}
\end{table}
\section{Conclusion and Future Work}
 \begin{table*}[t]
  \begin{center}
   \begin{tabular}{c|l} \hline \hline
\multicolumn{2}{c}{German$\rightarrow$Dutch example} \\ \hline
\textit{Zero 6L} & Een collega van mij had \textbf{toegang} tot \textbf{investeringsgegevens} van Fox guard \\ 
English meaning & A colleague of mine had \textbf{access} to \textbf{investment data} of Fox guard \\ \hline
\textit{Filtered Dict} & Een collega van mij had \textbf{Zugang} tot \textbf{investment} van de autoriteiten van Fox guard \\
English meaning & A colleague of mine had \textbf{Zugang} to \textbf{investment} from the authorities of Fox guard \\ \hline 
\textit{Reference} & Een collega van me kreeg \textbf{toegang} tot \textbf{investeringsgegevens} van Vanguard  \\
English meaning & A colleague of mine received \textbf{access} to \textbf{investment data} from Vanguard \\ \hline \hline 
\multicolumn{2}{c}{German$\rightarrow$Romanian example} \\ \hline
\textit{Zero 6L}  & Pentru că s-ar aștepta să apelăm la medic în \textbf{nächsten} dimineață . \\ 
English meaning & Because he would expect to call a doctor in \textbf{nächsten} morning . \\ \hline
\textit{Filtered Dict} & Pentru că s-ar aștepta să-l chemăm pe doctori în \textbf{următorul} dimineață . \\
English meaning & Because he would expect us to call the doctors the \textbf{next} morning . \\ \hline 
\textit{Reference} & Răspunsul e că cei care fac asta se așteaptă ca noi să ne sunăm doctorii în dimineața \textbf{următoare} .  \\
English meaning & The answer is that people who do this expect us to call our doctors the \textbf{following} morning . \\ \hline \hline 
   \end{tabular}
 \caption{\label{ExampleDict} Examples of the sentences with the words in wrong languages produced by \textit{Zero} systems and the corrected version 
 produced by the same systems having the target dictionary filtered in decoding phase. 
 Target Dictionary Filtering is not only helpful in producing readable and fluent outputs but also clearly affects to the choices of next words.}
  \end{center}
 \end{table*}
In this paper, we present our experiments toward zero-shot translation tasks using a multilingual Neural Machine Translation framework. 
We proposed two strategies which substantially improved the multilingual systems in terms of both performance and training resources.

On the future work, we would like to look closer to the outputs of the systems in order to analyze better the effects of our strategies. 
We also have the plan to expand our strategies on full multilingual systems, for more languages and different data conditions.

\section{Acknowledgements}
The project leading to this application has received funding from the European Union's Horizon 2020 research and innovation programme under grant agreement n$^\circ$ 645452.
The research by Thanh-Le Ha was supported by Ministry of Science, Research and the Arts Baden-W\"urttemberg.

\bibliographystyle{IEEEtran}
\bibliography{references}

\end{document}